\documentclass[9pt, conference]{IEEEtran}
\IEEEoverridecommandlockouts
\usepackage{cite}
\usepackage{amsmath,amssymb,amsfonts}
\usepackage{algorithmic}
\usepackage{graphicx}
\usepackage{textcomp}
\usepackage{xcolor}
\usepackage{amsmath}
\usepackage{amssymb}
\usepackage{bbm}
\usepackage{bm}
\usepackage[ruled,linesnumbered]{algorithm2e}

\usepackage{threeparttable}
\usepackage{graphicx}
\usepackage{subfigure}
\usepackage{multirow}
\usepackage{booktabs}
\usepackage{balance}
\usepackage{url}
\def\BibTeX{{\rm B\kern-.05em{\sc i\kern-.025em b}\kern-.08em
    T\kern-.1667em\lower.7ex\hbox{E}\kern-.125emX}}
\begin{document}

\title{On EDA-Driven Learning for SAT Solving}

% \author{Anonymous submission\vspace{10pt}}

% \author{
% Min Li, Zhengyuan Shi, Qiuxia Lai, Sadaf Khan, Shaowei Cai, Qiang Xu \\
% Department of Computer Science and Engineering \\
% The Chinese University of Hong Kong\\
% Shatin, N.T., Hong Kong\\
% \texttt{\{mli,zyshi21,qxlai,skhan,qxu\}@cse.cuhk.edu.hk}
% }

\author{\IEEEauthorblockN{Min Li\IEEEauthorrefmark{1}, Zhengyuan Shi\IEEEauthorrefmark{1}, Qiuxia Lai\IEEEauthorrefmark{1}, Sadaf Khan\IEEEauthorrefmark{1}, Shaowei Cai\IEEEauthorrefmark{2} and Qiang Xu\IEEEauthorrefmark{1}}
\IEEEauthorblockA{\IEEEauthorrefmark{1}\textit{Department of Computer Science and Engineering},
\textit{The Chinese University of Hong Kong},
Shatin, Hong Kong S.A.R.\\
\{mli,zyshi21,qxlai,skhan,qxu\}@cse.cuhk.edu.hk}

\IEEEauthorblockA{\IEEEauthorrefmark{2}\textit{Institute of Software}.
\textit{Chinese Academy of Sciences}, China \\
% \{wangliuzheng, wangnaixing, huangyu61\}@hisilicon.com
}
}

\maketitle

\begin{abstract}
We present \emph{DeepSAT}, a novel end-to-end learning framework for the Boolean satisfiability (SAT) problem. Unlike existing solutions trained on random SAT instances with relatively weak supervision, we propose applying the knowledge of the well-developed electronic design automation (EDA) field for SAT solving. Specifically, we first resort to logic synthesis algorithms to pre-process SAT instances into optimized and-inverter graphs (AIGs). By doing so, the distribution diversity among various SAT instances can be dramatically reduced, which facilitates improving the generalization capability of the learned model. Next, we regard the distribution of SAT solutions being a product of conditional Bernoulli distributions. Based on this observation, we approximate the SAT solving procedure with a conditional generative model, leveraging a novel directed acyclic graph neural network (DAGNN) with two polarity prototypes for conditional SAT modeling. To effectively train the generative model,  with the help of logic simulation tools, we obtain the probabilities of nodes in the AIG being logic `1' as rich supervision. We conduct comprehensive experiments on various SAT problems. Our results show that, DeepSAT achieves significant accuracy improvements over state-of-the-art learning-based SAT solutions, especially when generalized to SAT instances that are relatively large or with diverse distributions. 
\end{abstract}

% \vspace{-8pt}
\section{Introduction}
% \vspace{-4pt}

The Boolean satisfiability (SAT) problem, which determines whether a combination of binary input variables exists to satisfy a given Boolean formula, has a broad range of applications, such as planning~\cite{buttner2005satisfiability}, scheduling\cite{horbach2010boolean}, and verification~\cite{vizel2015boolean}. 

SAT is NP-complete. Over the past decades, many powerful heuristic-based SAT solvers are presented in the literature~\cite{sorensson2005minisat, fleury2020cadical}. 
%These manually-designed heuristics require significant insights into the problem.
% Consequently, a large amount of research has been devoted to SAT solving over the past several decades. 
% Recently, with the emergence of deep learning techniques, a fairly of studies investigate the problem of "\textit{learn to solve SAT}". Many methods aim at learning the heuristics inside the classical search-based algorithms~\cite{yolcu2019learning, kurin2020can, zhang2021nlocalsat}. While showing promising performance, the resulted model is bounded by the greedy strategy, which is sub-optimal in general~\cite{amizadeh2018learning}. As an alternative, a few works~\cite{selsam2018learning, amizadeh2018learning} learn to solve SAT from scratch in an end-to-end manner. To achieve this goal, a graph neural network is trained on random $k$-SAT instances to either predict SAT/UNSAT as a binary classification task, or generate a solution if the input circuit is indeed SAT. Meanwhile, to evaluate the generalization performance of the proposed methods, the well-trained model is tested on novel SAT distributions.
Recently, a family of data-driven techniques that apply deep learning (DL) for SAT solving has emerged. Some methods try to learn optimized search policies used in existing heuristic-based SAT solvers~\cite{kurin2020can,yolcu2019learning,zhang2021nlocalsat}. Though effective, the resulting model's performance is %bounded by the best solutions provided by classical heuristics.
%bounded by the best solutions provided by the general template defined by the search algorithms.
bounded by the best solutions provided by the general template of heuristics used in the underlying classical SAT solvers.
As an alternative, several end-to-end DL models are proposed for SAT solving~\cite{selsam2018learning, amizadeh2018learning,  duan2022augment}. These approaches try to learn SAT solutions from scratch, and they have the potential to achieve solutions beyond those of existing heuristic-based SAT solvers, despite the inferior performance as of now.

While promising, current end-to-end DL models for SAT solving share a few common weaknesses. On the one hand, 
% {
% \color{brown}
there exists notable distribution diversity among different SAT distributions, which makes generalization difficult.
% }
% using random SAT instances as training data 
% %obtained from various sources of SAT instances
% {
% \color{red}
% lacks a unified underlying distribution, significantly limiting learning efficiency and the generalization ability of the model.} 
% %For example, a SAT solver trained for the graph coloring problem may perform poorly for the vertex cover problem, since the distributions of the obtained training data differ significantly from each other (see Fig.~\ref{FIG:syn}). 
% % For example, NeuroSAT~\cite{selsam2018learning} directly uses the random SAT instances created with a manually-designed strategy for training, while DG-DAGRNN~\cite{amizadeh2018learning} utilize the circuit formats of the random SAT instances as training data. (not proper examples for demonstrating different 'distributions' of graph?)
On the other hand, the supervision cues used for model training are relatively weak. Existing models are either trained with binary classification labels (i.e., SAT or UNSAT)~\cite{selsam2018learning} or in an unsupervised manner~\cite{amizadeh2018learning}, resulting in unsatisfactory performance.

Given the above challenges, this work re-explores the essential elements of end-to-end DL models for SAT solving. We investigate it from a fundamentally different perspective: 
% {\color{red} we transfer the domain knowledge from the electronic design automation (EDA) field~\cite{jansen2003electronic} to SAT solving and reformulate the problem as a conditional modeling procedure.} 
we transfer the domain knowledge from the electronic design automation (EDA) field %~\cite{jansen2003electronic}
to SAT solving 
%and reformulate the problem as a conditional generative procedure, 
and mimic the Boolean constraint propagation (BCP)~\cite{wu2007qutesat} using a bidirectional propagation process.
% Specifically, we first apply logic synthesis algorithms to pre-process SAT instances into optimized and-invertor graphs (AIGs), a special form of circuit representation. Such pre-processing procedure maps different SAT distributions into a unified and compact distribution, which facilitates effective model learning.
% Second, we formulate the SAT solving as a generative modeling procedure following a joint Bernoulli distribution. To make the distribution tractable to model, we adopt a similar idea in PixelCNN~\cite{van2016pixel,oord2016conditional}, that is, factorize the joint Bernoulli distribution into a product of conditional univariate Bernoulli distributions of every variable. By applying efficient logic simulation on AIG circuits, we obtain a fairly accurate estimation of the parameter in univariate Bernoulli distributions, i.e., the probability of gates in the AIG being logic `1' as rich supervisions. A dedicated directed acyclic graph neural network (DAGNN) with two polarity prototypes is proposed to approximate the conditional probability, with only graph structure and conditions as inputs. Finally, we derive a simple solution sample scheme from the resulted conditional generative model to produce satisfying assignments. 

Specifically, we first propose a pre-processing procedure that applies logic synthesis algorithms~\cite{bjesse2004dag, mishchenko2006dag} to represent SAT instances as optimized and-invertor graphs (AIGs), thereby %{\color{red} generating a unified and compact distribution for training and testing.}
reducing the diversity between training and testing SAT distributions.
Second, we formulate the SAT solving process as a generative modeling procedure of the joint Bernoulli distribution of the binary inputs. Inspired by PixelCNN~\cite{oord2016conditional}, we factorize the joint Bernoulli distribution as a product of conditional univariate Bernoulli distribution of every variable. Next, we obtain rich supervision labels by applying efficient logic simulation on the AIG circuits to estimate the parameters in univariate Bernoulli distributions, i.e., the probability of signals in the AIG being logic `1'. 
The conditional generative model for approximating the conditional probability is realized as a dedicated directed acyclic graph neural network (DAGNN) with two polarity prototypes. The DAGNN learns a hidden space with good interpretability of logic values through a bidirectional propagation process conditioned on gate masks.
Finally, we produce satisfying assignments for the SAT instance from the trained conditional generative model using a simple solution sampling scheme.

% We refer to the proposed framework for end-to-end SAT solving as \textit{DeepSAT}. Overall, we make the following contributions: 1) To the best of our knowledge, DeepSAT is the first work that transfers the knowledge from EDA to learning-based SAT solving in a systematic way. In particular, we apply two well-established EDA techniques: logic synthesis and logic simulation, for data pre-processing and supervision construction. 2) We reformulate SAT solving as a generative modeling procedure, and hence leverage a directed acyclic graph neural network with two polarity prototypes to approximate the conditional distribution of SAT solutions. 3) The proposed DeepSAT achieves superior performance on both accuracy and generalization.
We refer to the proposed framework for end-to-end SAT solving as \textit{DeepSAT}. Overall, we make the following contributions: 
\begin{itemize}
    \item To the best of our knowledge, DeepSAT is the first work that systematically transfers the knowledge from EDA to learning-based SAT solving. In particular, we apply two well-established EDA techniques, namely logic synthesis and logic simulation, for reducing the distribution diversity of SAT instances and constructing the supervision labels, respectively.
    \item We reformulate the SAT solving process as a conditional generative modeling procedure to sequentially determine the value of the variables based on previously resolved variables, which enables the explicit SAT assignment prediction.
    \item We realize the conditional generative modeling using a novel DAGNN with two polarity prototypes, and propose to train DAGNN through bidirectional propagation conditioned on gate masks to mimic the BCP in traditional SAT solving. In this way, our DAGNN learns interpretable hidden states for solution sampling compared with other GNN-based solvers.
    
    % \item To the best of our knowledge, DeepSAT is the first work that systematically transfers the knowledge from EDA to learning-based SAT solving. In particular, we apply two well-established EDA techniques: logic synthesis and logic simulation, for data pre-processing and supervision construction.
    % \item We reformulate the SAT solving process as a conditional generative modeling procedure, which enables the explicit assignment prediction.
    % \item We propose a novel DAGNN with two polarity prototypes. Unlike the previous GNN-based SAT solvers that only use message passing scheme to enrich the node embeddings, the bidirectional propagation and polarity prototypes proposed in DeepSAT mimic the Boolean Constraint Propagation.
    % % \item {\color{blue}We show that the proposed model is a generalized version of Boolean constraint propagation (BCP), which enables solution sampling with polarity prototypes and bidirectional propagation.}
\end{itemize}
Experimental results show that the proposed DeepSAT solver achieves superior performance on both accuracy and generalization capabilities, compared to existing end-to-end learning-based solutions.

% \begin{itemize}

%     \item To the best of our knowledge, DeepSAT is the first work that transfers the knowledge from EDA to learning-based SAT solving in a systematic way. In particular, we apply two well-established EDA techniques: logic synthesis and logic simulation, for data pre-processing and supervision construction. 
    
%     \item We reformulate SAT solving as a generative modeling procedure, and hence leverage a directed acyclic graph neural network with two polarity prototypes to approximate the conditional distribution of SAT solutions.
    
%     \item We propose a simple yet efficient solution sampling scheme.
    
% \end{itemize}

% We organize the remainder of this paper as follows. We review related works in Section~\ref{sec:related}. Section~\ref{sec:model} introduces the DeepSAT architecture, while in Section~\ref{sec:sat-infer}, we present the conditional inference with a simple backtracking scheme to solve SAT. In Section~\ref{sec:exp}, we show experimental results on various SAT formulas. We discuss the limitations of proposed approach in Section~\ref{sec:discussion}. Finally, Section~\ref{sec:conclusion} concludes this paper.

% \vspace{-8pt}
\section{Related Work}\label{sec:related}
% \vspace{-4pt}
% Refer to our paper~\cite{li2021representation}.

\textbf{Learning-based SAT solvers.}
Applying deep learning techniques for combinatorial optimization (CO) has been explored in the last few years~\cite{khalil2017learning, hudson2021graph}. The problems of interest are often NP-complete, and traditional methods rely on heuristics to produce approximate solutions for large problem instances~\cite{yolcu2019learning}. Among them, as one of the most fundamental problems, SAT has become a popular target for learning-based solutions. 

There are mainly two kinds of learning-based SAT solvers in the literature. Some propose to use neural networks to learn the optimal heuristics within the conventional SAT solvers automatically~\cite{yolcu2019learning, kurin2020can}. Nevertheless, the performance of these methods is bounded by the heuristic-based framework, which is sub-optimal in nature~\cite{amizadeh2018learning}. 
Alternatively, we could train a deep learning model towards solving SAT from scratch~\cite{selsam2018learning, amizadeh2018learning}. The representative approaches include NeuroSAT~\cite{selsam2018learning} and DG-DAGRNN~\cite{amizadeh2018learning}. Specifically, NeuroSAT processes SAT problems as a binary classification task and proposes a clustering-based post-processing analysis to find SAT solutions. 
DG-DAGRNN approximates logic calculation with an evaluator network and trains the model to maximize the \textit{satisfiability value} for SAT instances. % {\color{red}The proposed method in this paper also belongs to this line, yet differs from works in the literature significantly: we re-formulate SAT solving as a generative modeling procedure and utilize bidirectional propagation with polarity prototypes for condition propagation.}

% {
% \color{red}
% New learning-based SAT solvers we should consider to refer: \cite{duan2022augment, shi2021transformer, zhang2022towards, garzon2022performance, wang2021heuristic, ozolins2021goal, wang2021neurocomb, guo2022machine, satorras2021neural, kuck2020belief, kyrillidis2020fouriersat, sun2022generalized, azar2020nngsat, chen2019graph,  geisler2021generalization}
% }

% \vspace{-10pt}
\textbf{Graph neural networks.}
Graph neural networks (GNNs) effectively model non-Euclidean structured data and achieve impressive performances for various challenging problems, thereby attracting lots of attention from both academia and industry~\cite{wu2020comprehensive}. 
%For example, \cite{zhang2019d} proposes a variational autoencoder for DAGs and apply it to neural architecture search and Bayesian network structure learning. 
Circuits can be naturally modeled as directed acyclic graphs (DAGs) with logic relationships. Recently, several GNN techniques are proposed to handle DAGs~\cite{thost2021directed}, and there are also a few DAG-GNN solutions dedicated for circuit analysis~\cite{li2021representation}. 

%In this work, we leverage directed acyclic GNN~\cite{zhang2019d, amizadeh2018learning, thost2021directed} as the basic machinery to capture computational dependencies of a Boolean circuit. 

% \paragraph{Electronic design automation.}

% \vspace{-8pt}
\section{DeepSAT}\label{sec:model}
% \vspace{-4pt}

In this section, we introduce the proposed DeepSAT framework in detail. To facilitate understanding, we first show preliminaries of SAT solving problem in Section~\ref{subsec:preliminaries}, where three different representations of SAT instances are presented. In Section~\ref{subsec:transformation}, we introduce our logic synthesis based pre-processing procedure. In Section~\ref{subesec:task}, we present our formulation of SAT solving as a conditional generative procedure and the construction of the supervision labels. We detail the model design in Section~\ref{subsec:model}, and show how to produce satisfying assignments with the trained conditional generative model in Section~\ref{subsec:sat-infer}.

% \begin{figure}[t!]
% 	\centering
% 	\includegraphics[width=0.7\linewidth]{figs/circuit_rep.pdf}
% 	\vspace{-10pt}
% 	\caption{Three graph representations of the Boolean formula in Equation~\ref{eq:sat_example}.}
% 	\label{FIG:graphRep}
% 	\vspace{-15pt}
% \end{figure}
\vspace{-4pt}
\subsection{Preliminaries}\label{subsec:preliminaries}
% \vspace{-4pt}
A Boolean logic/propositional formula $\phi$ takes a set of $I$ variables $\{x_i\}^I_{i=1}\!\in\!\{True, False\}$, and combines them using Boolean operators $\{\text{AND}\!-\!\land, \text{OR}\!-\!\lor, \text{NOT}\!-\!\neg, \cdots \}$, returning logic `$1$' (\textit{True}) or  logic `$0$' (\textit{False}). The Boolean satisfiability (SAT) problem asks whether there exists at least one combination of binary input $\{x_i\}^I_{i=1}$ for which a Boolean logic formula $\phi$ returns logic `1'. If so, $\phi$ is called \textit{satisfiable}; otherwise \textit{unsatisfiable}.

% {
% \color{red}\textbf{consider simplify this part.}
% }

There are various forms for representing a Boolean formula. The most commonly used form is \textit{conjunctive normal form} (CNF)%~\cite{tseitin1983complexity}
, wherein the variables are organized as a conjunction ($\land$) of disjunctions ($\lor$) of variables. % An example of CNF formula with three variables is shown in Equation~\ref{eq:sat_example}. 
By convention, each disjunction inside parentheses is termed a \textit{clause}, and each variable (possibly negation) within a clause is termed a \textit{literal}. Such CNF forms are used in NeuroSAT~\cite{selsam2018learning} and its follow-ups~\cite{yolcu2019learning, kurin2020can, zhang2021nlocalsat, duan2022augment} to represent Boolean formulas, resulting in %undirected 
bipartite graphs with two node types: literal and clause. 
Besides CNF, a Boolean formula can be represented in a \textit{Boolean circuit} format (also known as Circuit-SAT), wherein the primary inputs (PIs) of the circuit denote variables of a Boolean formula, and the internal gates denote Boolean operators. In this way, more structure information can be embedded in the inputs. A usage of Circuit-SAT can be found in~\cite{amizadeh2018learning}, where the CNF instances constructed from Boolean formulas are further converted to circuits, and encoded as directed acyclic graphs (DAGs). % See the CNF-converted circuit in Figure~\ref{FIG:graphRep}(b).

In this work, inspired by the common practice in many EDA processes, we propose to represent Boolean formulas in a unique format of circuits, i.e., \textit{and-invertor graph} (AIG) which is more brief compared with normal circuits. An AIG contains three types of nodes: PIs, two-input $\text{AND}$ gates, and one-input $\text{NOT}$ gates. %(see Figure~\ref{FIG:graphRep}(c)). 
It is theoretically guaranteed that any Boolean circuit has an AIG counterpart. By representing SAT instances in AIGs, we can bridge SAT solving and advanced EDA algorithms, e.g., using logic synthesis to pre-process the circuit forms (See Section~\ref{subsec:transformation}).

\begin{figure*}[t!]
	\centering
	\includegraphics[width=0.6\linewidth]{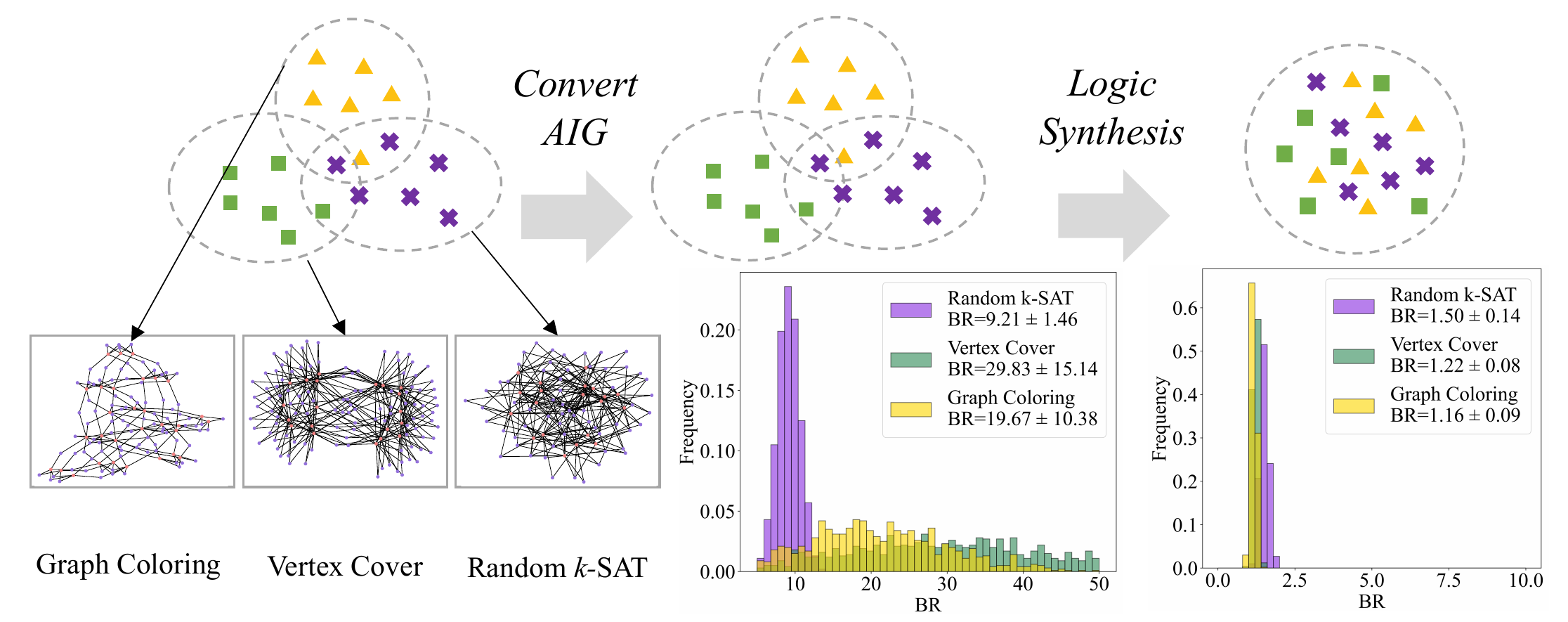}
	\vspace{-12pt}
	\caption{Pre-processing via logic synthesis.}
	\label{FIG:syn}
	\vspace{-10pt}
\end{figure*}

\vspace{-4pt}
\subsection{Pre-processing via Logic Synthesis}\label{subsec:transformation}
% \vspace{-4pt}

\iffalse
Though having CNFs as a common input format, previous learning based SAT solvers~\cite{selsam2018learning, yolcu2019learning, kurin2020can, zhang2021nlocalsat} showing poor generalization ability for different SAT classes, i.e., they need to design different heuristics for input SAT formulas of different distributions. Examples of the distribution diversity is visualized in Figure~\ref{FIG:syn}, where three SAT instances derived from different SAT classes, namely \textit{Random k-SAT}, \textit{Vertex Cover}, and \textit{Graph Coloring}, follow different distributions regarding graph structures. 
Yolcu \textit{et al.}~\cite{yolcu2019learning} also observe that the performance of heuristic specialized to a particular SAT distribution degrades considerably on other distributions. 
\fi

%Though having CNFs as a common input format, previous
Existing learning based SAT solvers~\cite{selsam2018learning, yolcu2019learning, kurin2020can, zhang2021nlocalsat} show poor generalization ability for different SAT classes, i.e., they need to design different heuristics for %input SAT formulas of 
different distributions of SAT problems. Figure~\ref{FIG:syn} shows the distribution diversity among three SAT instances, each belonging to a different SAT class. %Examples of the distribution diversity is visualized in Figure~\ref{FIG:syn}, where three SAT instances derived from different SAT classes, namely \textit{Random k-SAT}, \textit{Vertex Cover}, and \textit{Graph Coloring}, follow different distributions regarding graph structures. 
\cite{yolcu2019learning} also observe that the performance of heuristic specialized to a particular SAT distribution degrades considerably on other distributions. 

\iffalse
Instead of learning different models for different distributions, we leverage the \textit{logic synthesis} techniques~\cite{cortadella2003timing, bjesse2004dag, mishchenko2006dag} in the EDA field to reduce the distribution diversity among different SAT distributions to facilitate SAT solving. In this way, the model trained on one SAT source can generalize well to test data from different SAT sources. 
% An illustration of the distribution mapping based on logic synthesis is shown in Figure~\ref{FIG:syn}.
Specifically, we first represent the given SAT instances in AIG format. Then we apply two techniques of logic synthesis, namely logic rewriting~\cite{bjesse2004dag, mishchenko2006dag} and logic balancing~\cite{cortadella2003timing}, to reduce the distribution diversity among SAT instances from different sources. %As demonstrated in Figure~\ref{FIG:rwbl}(a),
Logic rewriting can reduce the total number of AIG nodes, and logic balancing  further produces a more balanced graph structure by constructing an equivalent circuit with minimal logic levels. %(Figure~\ref{FIG:rwbl}(b)). 
Please refer to~\cite{brayton2010abc, een2007applying} and Appendix~\ref{appendix:syn} for more technical details about logic synthesis techniques.
\fi
To this end, we leverage the \textit{logic synthesis} techniques~\cite{bjesse2004dag, mishchenko2006dag} in the EDA field to reduce the distribution diversity among different SAT distributions, thus improving the generalizability of DeepSAT.  %In this way, the model trained on one SAT source can generalize well to test data from different SAT sources. 
% An illustration of the distribution mapping based on logic synthesis is shown in Figure~\ref{FIG:syn}.
Specifically, we first represent the given SAT instances in AIG format. Then we apply two logic synthesis techniques, i.e., logic rewriting~\cite{ mishchenko2006dag} and logic balancing~\cite{cortadella2003timing}. %to reduce the distribution diversity among SAT instances from different sources. %As demonstrated in Figure~\ref{FIG:rwbl}(a),
The former reduces the total number of AIG nodes, and latter  constructs a more balanced circuit with minimal logic level. %thus producing a more balanced graph. %structure by constructing an equivalent circuit with minimal logic levels. %(Figure~\ref{FIG:rwbl}(b)). 
Please refer to~\cite{brayton2010abc} %and Appendix~\ref{appendix:syn} 
for more technical details about these techniques.

\iffalse
To quantitatively show the change of the distributions before and after logical synthesis, we further introduce a scale-independent measurement, i.e., balance ratio (BR) ~\cite{walker1976locally}, and represent the AIG distributions using the frequency histogram of BR values. 
BR is defined as the average ratio of larger fanin region size to smaller fanin region size for each two-fanin gate, i.e. AND gate in AIG, thus can reflect the balance degree of the binary fanin regions. A BR value closer to 1 indicates more balanced fanin regions of the gate. 
As shown in Figure~\ref{FIG:syn}, AIGs from different SAT sources %, namely \textit{Random k-SAT}, \textit{Vertex Cover}, and \textit{Graph Coloring}, 
have distinct frequency histogram regarding BR values. The three frequency histograms become similar after performing logic synthesis, where all the AIGs are optimized to have more balanced fanin regions of the gate, thus having BR values closer to 1. 
This clearly demonstrate that logic synthesis can reduce the distribution diversities among AIGs from different SAT sources, which naturally introduces a strong inductive bias (i.e., compactness and balance) of AIG-based SAT instances circuits for more effective SAT solving. 
\fi
We quantitatively show the change of the distributions before and after logical synthesis, using a scale-independent measurement, i.e., balance ratio (BR)~\cite{walker1976locally} - the average ratio of larger fanin region size to smaller fanin region size for each two-fanin gate, i.e., AND gate in AIG. %and represent the AIG distributions using the frequency histogram of BR values. 
%It is defined as the average ratio of larger fanin region size to smaller fanin region size for each two-fanin gate, i.e. AND gate in AIG. %thus can reflect the balance degree of the binary fanin regions. 
A BR value closer to 1 indicates more balanced fanin regions of the gate. 
As shown in Figure~\ref{FIG:syn}, AIGs from different SAT sources %, namely \textit{Random k-SAT}, \textit{Vertex Cover}, and \textit{Graph Coloring}, 
have distinct frequency histogram of BR values. But after performing logic synthesis, these histograms become similar showing BR values close to 1, since all the AIGs are optimized to have more balanced fanin regions of the gate. % thus having BR values closer to 1. 
This demonstrates that logic synthesis can reduce the distribution diversities among AIGs from different SAT sources, which naturally introduces a strong inductive bias (i.e., compactness and balance) %of AIG-based SAT instances circuits
for more effective SAT solving. 
% \vspace{-8pt}
\subsection{SAT Solving as a Conditional Generative Procedure}\label{subesec:task}
% \vspace{-4pt}

\textbf{Problem formulation.} As introduced in Section~\ref{subsec:preliminaries}, we represent a Boolean formula as an AIG-based directed graph $\mathcal{G}= (\mathcal{V}=\{\mathcal{V}_P,\mathcal{V}_G\}, \mathcal{E})$, where the nodes $\mathcal{V}_P = \{v_i\}^I_{i=1}$ correspond to the primary inputs (PIs), the nodes $\mathcal{V}_\mathcal{G}\!=\!\{v_i\}_{i=I+1}^{|\mathcal{V}|}$ denote the logic gates, and the directed edges $\mathcal{E}$ represent wires among the gates.
% , and solve SAT problem by approximating the joint distribution of the primary inputs (PIs) and the distribution of the primary output (PO) using a deep learning model. 
The PI states can be viewed as a random variable following the multivariate Bernoulli distribution, $\mathbf{x}\!=\!(x_1, x_2, \cdots, x_I)^T\!\in\!\{0, 1\}^I\!\sim\!\text{Bernoulli}(\bm{\theta})$, where $I$ is the number of PIs and $\bm{\theta}$ is the parameter of the distribution. 
The primary output (PO) $y$ is the state of the gate at the last logic level, which follows the univariate Bernoulli distribution, indicating whether the Boolean circuit evaluates to logic `1' or not. 
A feasible solution for a satisfiable Boolean circuit can be derived by:
\begin{equation}
   \text{Solution } \mathbf{x}^* = \arg\max_{\mathbf{x}} p(\mathbf{x} |y=1;\mathcal{G},\bm{\theta}).
   \label{eq:infer_target}
\end{equation}
Similar to PixelCNN~\cite{oord2016conditional}, we handle the intractable joint probability distribution of $\mathbf{x}$ by factorizing it into a product of conditional univariate Bernoulli distributions:
\begin{equation}
    \small
	p(\mathbf{x} |y=1;\mathcal{G},\bm{\theta}) = \prod_{i=1}^{I}p(x_i|\mathbf{x}_{<i}, y=1; \mathcal{G},\theta_i), ~~x_i|\mathbf{x}_{<i}\sim \text{Bernoulli}(\theta_i),
    \label{eq:conditional}
\end{equation}
where $\mathbf{x}_{<i}\!=\!\{x_1, x_2, \cdots, x_{i-1}\}$ denote the PIs of which the values have been determined, and $p(x_i|\mathbf{x}_{<i}, y=1; \mathcal{G},\theta_i)\!=\!\theta_i^{x_i}(1-\theta_i)^{x_i}$. 

\textbf{Supervision label construction.} 
To obtain the joint probability in Equation~\ref{eq:conditional}, we need to figure out $\theta_i$, and then calculate $p(x_i|\mathbf{x}_{<i})$. 
Note that different from PixelCNN~\cite{oord2016conditional}, we do not have a natural fixed order for picking $x_i$s. 
To facilitate subsequent discussions, we rewrite the condition terms $\mathbf{x}_{<i}$ in Equation~\ref{eq:conditional} as $\mathbf{x}_m$. The determined PIs are designated using a mask $\mathbf{m}= (m_1, \cdots, m_{|\mathcal{V}|})^T \in \{1, 0, -1\}^{|\mathcal{V}|}$ to simulate the generative procedure, which is applied on all nodes $\mathcal{V}$ to indicate whether the state of a node is determined or not:
\begin{equation}
    m_j=\left\{
    \begin{array}{lcl}
        0       &      & \text{if}\quad v_j\in\mathcal{V}_G~\text{or}~x_j\notin\mathbf{x}_m\\
        1       &      & \text{if}\quad x_j\in\mathbf{x}_m~\text{and}~x_j=\textit{True,}\\
        -1      &      & \text{if}\quad x_j\in\mathbf{x}_m~\text{and}~x_j=\textit{False.}
    \end{array}
    \right.
    \label{eq:mask}
\end{equation}

% To construct supervision labels $\hat{\theta_i}$, 
Following a frequentest setting, we estimate a value for each $\theta_i$ by maximum likelihood estimation (MLE). %~\cite{bishop2006pattern}.
Suppose we feed $N$ random assignments (also known as \textit{logic simulation} in EDA context) to the Boolean circuit $\mathcal{G}$ and observe variable $x_i$ being logic `1' for $M$ times. Then we can obtain the maximum likelihood estimator for $\theta_i$ of $p(x_i; \mathcal{G},\theta_i)$:
\begin{equation}
    \hat{\theta_i} = \frac{M}{N}.
\end{equation}
It should be noted that such estimation is achieved without any conditions, e.g., the circuit being satisfiable or values of some nodes are known in advance. If some conditions are imposed into the circuit, we simply filter out the random assignments that violate the conditions during logic simulation, therefore can estimate $\theta_i$ in the conditional distribution. 

% {
% \color{red}
% All-SAT problems. Organize the rebuttal as part of appendix.
% }

We refer $\hat{\theta_i}$ as the \textit{simulated probability} of node being logic `1' in the following discussion. To avoid in-accurate estimation associated with maximum likelihood, we perform logic simulations with a large number of random patterns (e.g., $15$k random patterns to each AIG in our experiments) to obtain faithfully simulated probability values. Alternatively, for larger problems, a practical way is to first use an efficient \textit{all solutions SAT solver}~\cite{toda2016implementing} to obtain all possible satisfying solutions, and then estimate the supervision signal $\hat{\theta}_i$ from these assignments.
In general, the cost of our logic simulation is on par with those of other SAT solving algorithms such as NeuroSAT~\cite{selsam2018learning}, where the SAT-related supervision labels are generated from a classical SAT solver. % Please refer to Appendix~\ref{appendix:simulation} for more details about dataset construction.

% Furthermore, we construct random k-SAT instances with a low ratio of clauses to variables (C/V ratio), so that the total number of satisfied assignments is relatively large.

\textbf{Training objective.}
The core of our approach is a graph-based model taking a DAG $\mathcal{G}$ and the conditions $\mathbf{m}$ as inputs to approximate the simulated probabilities at the node-level directly. Our hypothesis is that a DAG structure sufficiently defines the logic computation of the Boolean circuit and characterizes the dependence of the different gates/variables. With enough relational inductive biases, a parameterized graph-based model is able to predict the simulated probability of all gates under logic simulation. 
Driven by this hypothesis, we designate the training objective as the following mapping:
\begin{multline}
         \mathcal{F}: (\mathcal{G}, \mathbf{m}) \mapsto  \hat{\bm{\theta}},\\ \text{ where} \quad
     \hat{\bm{\theta}} = \{ \hat{\theta}_i = \arg \max_{\theta_i} p(x_i|\mathbf{x_m}, y=1; \mathcal{G}, \theta_i)  | x_i \in \mathcal{G} \},
     \label{eq:training_obj}
\end{multline}

where the condition $\mathbf{x_m}$ corresponds to the determined nodes defined by the mask $\mathbf{m}$ (Equation~\ref{eq:mask}). 
% % Specifically, we approach satisfiability by \textit{Masked Simulated Probability Prediction} (MSPP). 
% In other words, we randomly mask some of the gates with random binary logic values (i.e., either logic `1' or logic `0') defined by $\mathbf{m}$, and the objective is to regress the simulated probabilities of the remaining gates being logic `0'. Each conditional probability is parameterized by a deep neural network whose architecture is chosen according to the required inductive biases for the graph-structured Boolean circuits. To learn the function $\mathcal{F}$, we use the least absolute error loss (i.e., L1 loss) that penalizes errors in prediction. The model architecture for $\mathcal{F}$ is elaborated in Sec.~\ref{subsec:model}.
We generate a mask $\mathbf{m}$ initialized from a zero vector by fixing the element for PO to 1 (i.e., let $y\!=\!1$), and then assign the elements corresponds to a randomly designated condition terms $\mathbf{x}_m$ with 1 or -1. Given $\mathbf{m}$, the objective is to regress the simulated probabilities of the remaining nodes $v_i$ being logic `1', i.e., to approximate $\hat{\theta_i}$. Each conditional probability is parameterized by a deep neural network whose architecture is chosen according to the required inductive biases for the graph-structured Boolean circuits. The function $\mathcal{F}$ is learned by minimizing the least absolute error between the prediction and the supervision label $\hat{\bm{\theta}}$. % The model architecture for $\mathcal{F}$ is elaborated in Section~\ref{subsec:model}.

% The constructed training dataset $\mathcal{D}$ consists of tuples of $(\mathcal{G}^{(i)}, \mathbf{m}^{(i)}, {\hat{\bm{\theta}}}^{(i)})$, where $\mathcal{G}^{(i)}$ is a Boolean circuit, $\mathbf{m}^{(i)}$ is a conditional mask and ${\hat{\bm{\theta}}}^{(i)}$ is the simulated probabilities for every node in the Boolean circuit. 
% In practice, we also include training samples where we consider $y\!=\!0$ when generating the mask $\mathbf{m}$. 
% Please refer to Appendix~\ref{appendix:simulation} for more details about training dataset construction.

\begin{figure}[t!]
	\centering
	\includegraphics[width=\linewidth]{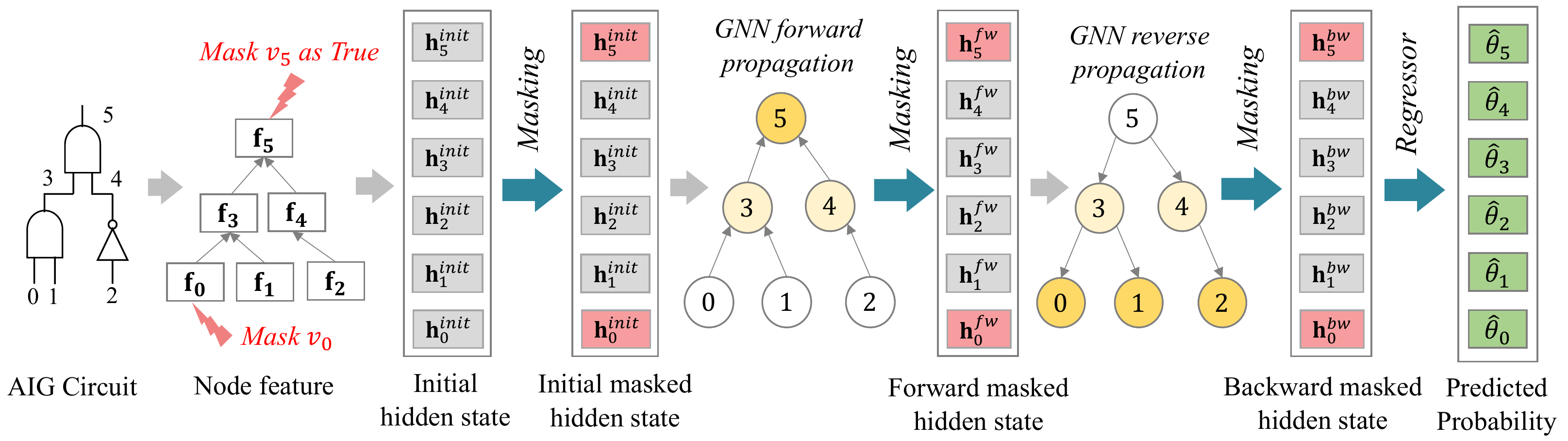}
	\vspace{-5pt}
	\caption{The overview of model architecture. We use polarity prototypes to replace the hidden states of gates with masks during forward and reverse propagation.}
	\label{FIG:model}
	\vspace{-10pt}
\end{figure}

\vspace{-4pt}
\subsection{The Proposed Model}\label{subsec:model}
% \vspace{-4pt}

% Given a circuit graph $\mathcal{G}$ in AIG form and its corresponding mask $\mathbf{m}$, the objective of our model is to estimate the simulated probability $\hat{\bm{\theta}}$ for all nodes in the circuit. 
To capture directed relational bias embedded in circuits, we leverage the dedicated directed acyclic GNNs (DAGNNs)~\cite{thost2021directed} as the model $\mathcal{F}$ in Equation~\ref{eq:training_obj} to learn structural information and the computational behavior of logic circuits, and embed them as vectors on every logic gate~\cite{li2021representation}. As shown in Figure~\ref{FIG:model}, the model consists of two stages of directed acyclic GNN propagation along with the masking operations defined in Equation~\ref{eq:mask}. Then a multi-layer perceptron (MLP) regressor is applied on node vectors for predicting the simulated probability. % In Appendix~\ref{appendix:gnn}, different GNN designs are investigated.

Given a circuit graph $\mathcal{G}$, we use $\mathbf{f}_i$ and $\mathbf{h}_i$ to denote the gate type and the hidden vector, respectively. Specifically, as three types of gates, i.e., PI, AND and NOT, are present in AIGs, we encode the gate type as a $3$-d one-hot embedding for each node according to its gate type. %The hidden vector $\mathbf{h}_i$ is initialized randomly by a standard Gaussian distribution. 
We elaborate the details of the model design in DeepSAT below.

\textbf{Polarity prototypes.}
The training objective derived in Equation~\ref{eq:training_obj} requires the model to be able to condition on some given gate states. %For example, when the first PI is masked as logic $1$, the model shall approximate the following distribution: $p(\mathbf{x}_{i>1}|x_1=1)$. 
To enable such conditional modeling, 
% , we shall encode the conditions to some specific form that the model can accept. Specifically, 
we define two prototypes~\cite{snell2017prototypical},  $\mathbf{h}_{pos}$ and $\mathbf{h}_{neg}$, for two polarities of node probabilities, one for node being logic `1' (positive polar) and the other for node being logic `0' (negative polar), respectively. 

Assuming every node $v_i$ is encoded as a hidden vector $\mathbf{h}_i$, the hidden vectors for the masked nodes are replaced by their corresponding prototypes according to the value of the mask:
% ($\mathbf{h}_i^\prime = \mathbf{h}_{pos}$ in the above example) as follows:
\begin{equation}
    \mathbf{h}_i^\prime=\left\{
    \begin{array}{lcl}
        \mathbf{h}_{pos}       &      & \text{if}\quad{m_i = 1,}\\
        \mathbf{h}_i       &      & \text{if}\quad{m_i = 0,}\\
        \mathbf{h}_{neg}       &      & \text{if}\quad{m_i = -1.}
    \end{array}
    \right.
    \label{eq:hidden_mask}
\end{equation}
In practice, the initial hidden vectors are sampled from a normal Gaussian distribution. Therefore, we fix the hidden vectors for two polarities as two near-boundary points, i.e, all positive ones for logic $1$ ($\mathbf{h}_{pos} = [1, 1, \cdots. 1]$) and all negative ones for logic $0$ ($\mathbf{h}_{neg}=[-1, -1, \cdots, -1]$), respectively. The two polarity prototypes can be thought of as two extreme points lying on opposite sides of a sphere in hidden space. Intuitively, during training, the hidden vectors of  gates with a probability close to $1$ will be pulled to all ones, and vice versa. Therefore, the two polarity prototypes facilitate learning a continuous and compact hidden space with good \textit{interpretability} of logic values. Note that other polarity prototypes are also feasible, as long as the two polarity prototypes are far away from each other in the hidden space. % Detailed explanation of the polarity prototypes are included in Appendix~\ref{appendix:pp}% We have also tried learnable prototypes similar to learnable tokens in the Transformer-based models~\cite{dosovitskiy2021an, he2021masked, devlin2018bert}, but do not observe performance gain (Appendix~\ref{appendix:learnable_prototypes}).

For notational simplicity, we omit the superscript prime '$^\prime$', and directly use $\mathbf{h}_i$ to denote the input hidden vector after the masking operation defined in Equation~\ref{eq:hidden_mask}.
% assume the input hidden vectors $\mathbf{h}_i$ of GNN functions are processed by the masking operation defined in Equation~\ref{eq:hidden_mask}. 

\textbf{Forward propagation.} To aggregate the information from node $v$'s predecessors, we implement the attention mechanism in the additive form~\cite{thost2021directed} as:
\begin{multline}
    \mathbf{a}_v^{fw} = \sum_{u \in \mathcal{P}(v)}  \alpha_{uv}^{fw}  \mathbf{h}_u^{init}, \\
\text{\quad where\quad} \alpha_{uv}^{fw} = \mathop{softmax}\limits_{u \in \mathcal{P}(v)} (w_1^\top \mathbf{h}_v^{init}+ w_2^\top \mathbf{h}_u^{init}),
\label{eq:attn}
\end{multline}
where $\mathcal{P}(v)$ denotes the set of direct predecessors of $v$. Using the `Attention' language, the inital hidden state $\mathbf{h}_v^{init}$ serves as a \textit{query}, and the representation of predecessors $\mathbf{h}_u^{init}$ serves as a \textit{key}. 
We then use the Gated Recurrent Unit (GRU) to combine the aggregated information with $v$'s information, including the encoded gate type $\mathbf{f}_v$ and the initial hidden vector $\mathbf{h}_v^{init}$ to update the hidden vector of target node $v$:
\begin{equation}
    \mathbf{h}_v^{fw} = GRU([\mathbf{a}_v^{fw},\mathbf{f}_v], \mathbf{h}_v^{init}),
\end{equation}
wherein $\mathbf{a}_v^{fw}$, $\mathbf{f}_v$ are concatenated together and treated as input, while $\mathbf{h}_v^{init}$ is the past state of GRU. 
The above functions process nodes according to the logic levels defined by circuit's topological orders, starting from PIs and ending up with the single PO. After that, the node hidden vector $\mathbf{h}_v^{fw}$ is further updated by according to the same mask following Equation~\ref{eq:hidden_mask}.% to ensure that the polarity regularization described in Sec.~\ref{subsec:regularization} is still maintained in node's hidden space.

% On the other hand, our proposed GNN model differs from previous DAG-GNNs~\cite{amizadeh2018learning, zhang2019d, thost2021directed} that initialize $\mathbf{h}_v^0$ as $\mathbf{x}_v$ and treat the aggregated message as the state of recurrent function. In contrast, we fix the gate type information of nodes $\mathbf{x}_v$ as inputs for all iterations. Such employment can avoid the information vanishing of gate properties during the long-term recursive propagation. 

\textbf{Reverse propagation.}
To model the effect of the conditional term related to satisfiability, i.e., $y\!=\!1$ in Equation~\ref{eq:training_obj}, we perform a reverse information propagation after the forward propagation.
Specifically, we impose the condition of satisfiability PO by masking its hidden vector as $\mathbf{h}_{pos}$, and process the graph in reversed topological order to propagate the condition to PIs. 
% We also consider backward information propagation, i.e., processing the graph in reversed topological order, %\footnote{The backward propagation differs from the concept of gradient backward propagation in deep learning training.}. 
% which imposes the condition of satisfiability (by masking the hidden vector of PO as $\mathbf{h}_{pos}$) from PO to PIs. 
By doing so, the updated node hidden vector $\mathbf{h}_v^{bw}$ contains extra information about the satisfiability compared with $\mathbf{h}_v^{fw}$ that is obtained only using forward propagation.
% is conditioned on satisfiability and hence can model the conditional probability in Equation~\ref{eq:training_obj}.

The backward propagation is similar to the forward propagation regarding computation of aggregation and combination function, except that during the aggregation, we only consider the direct successors $\mathcal{S}(v)$ of target node $v$. 
The hidden vector $\mathbf{h}_v^{bw}$ is further processed using the same mask as above following Equation~\ref{eq:hidden_mask}, and fed into the regressor for predicting the simulated probabilities.

\begin{figure}[t!]
	\centering
	\includegraphics[width=\linewidth]{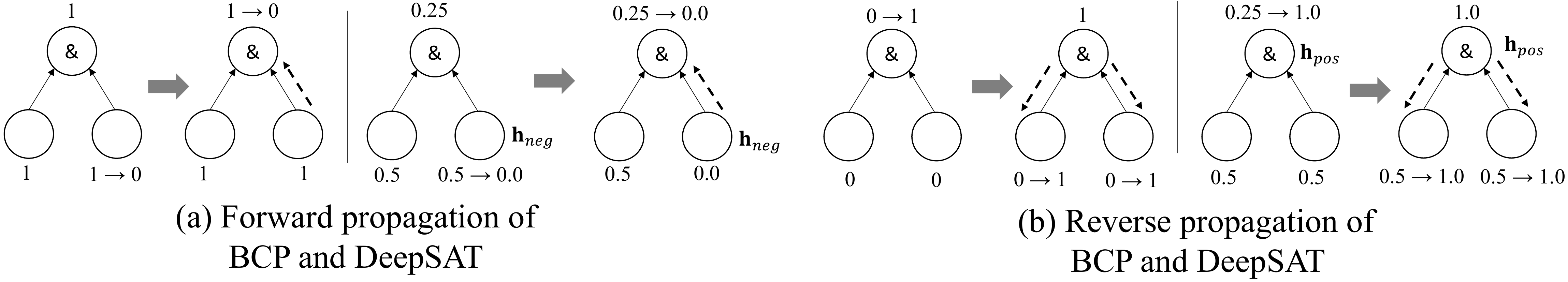}
	\vspace{-5pt}
	\caption{Example of forward/reverse propagation in BCP and DeepSAT. DeepSAT enables constraint propagation by bidirectional propagation and polarity prototypes during learning.}
	\label{fig:bcp}
	\vspace{-10pt}
\end{figure}

\textbf{Comparison with previous methods using bidirectional propagation.}
Adopting a reverse propagation besides the forward propagation for training DAG structures is not new in GNN-based SAT solving~\cite{amizadeh2018learning, thost2021directed}, yet we highlight key differences that support the superior SAT solving performance of DeepSAT. 
In DAGNN~\cite{thost2021directed} and DG-DAGRNN~\cite{amizadeh2018learning}, the reason to perform forward and reverse propagation is to enrich the node state vectors with information from both the ancestor and the descendant nodes. While in DeepSAT, the goal of using bidirectional propagation along with polarity prototypes is to mimic the Boolean constraint propagation (BCP)~\cite{belov2010improved, zhang2021circuit} mechanism in logic reasoning.

Specifically, BCP is initiated when a logic value is assigned to an unassigned gate $v$, which leads to value changing of some unassigned gates in the neighbourhood of $v$ (i.e., fanin and fanout). There are two types of propagation considering the direction: forward propagation and reverse propagation (Figure~\ref{fig:bcp}). Similarly, in DeepSAT, we introduce polarity prototypes as hidden states for gates assigned with known logic states. Specially, the satisfiability is modeled by assigning logic `1' to PO. To this end, optimizing the conditional gate-level probability $ p(x_i|\mathbf{x_m}, y=1; \mathcal{G}, \theta_i)$ (Equation~\ref{eq:conditional}) with \textit{bidirectional propagation and polarity prototypes} enables learning BCP in the hidden space (Figure~\ref{fig:bcp}). Such design encourages a continuous and compact hidden space with good \textit{interpretability} of logic values. More importantly, with polarity-regularized hidden space, we can manipulate the states of nodes and thus perform conditional solution sampling as described below.

% \vspace{-8pt}
\subsection{Solution sampling scheme} \label{subsec:sat-infer}
% \vspace{-4pt}

To sample solutions from the well-trained conditional model, we iteratively select the undetermined PI with the highest \textit{confidence} values predicted by the model, i.e., the PI with probability prediction closest to $0$ or $1$. Specifically, given an SAT instance $\mathcal{G}$ with $I$ variables, we conduct the following \textit{auto-regressive} procedure: 1) We mask PO as logic `1', and generate the corresponding mask vector $\mathbf{m}_0$. 2) At iteration $t$, we pass $(\mathcal{G},\mathbf{m}_{t})$ to well-trained DeepSAT model to estimate the simulated probability of all un-masked PIs. The PI with the highest confidence is selected and masked as logic `0' if the prediction is smaller than 0.5, otherwise logic `1'. According to the selected PI and its masked value, a new mask $\mathbf{m}_{t+1}$ is generated. 3) Repeat the second step until all PIs have been masked ($I$ iterations in total). Then we obtain a candidate assignments $\hat{\mathbf{x}}$ from $\mathbf{m}_{I}$.

To explore more possible assignments to satisfy the SAT instance, we also develop a simple \textit{flipping}-based strategy to sample more solutions from the conditional model ($(I+1)$ solutions in the worst case). Briefly, if the initial sampled assignment is not the satisfying solution, we record the order of PIs being masked and flip a specific PI in the next sampling step, following the recorded order. % The implementation details, as well as the pseudo-code, are provided in Appendix~\ref{appendix:infer}.

% {\color{red} Unlike previous recurrent models (i.e., NeuroSAT~\cite{selsam2018learning} and DG-DAGRNN~\cite{amizadeh2018learning}) that require a large number of propagation iterations to find a solution, our DeepSAT model generates a solution in each sampling procedure.

% \textbf{Explain this part. Very important.}
% }

% \subsection{Comparison between DeepSAT and DG-DAGRNN}
% {
% \color{red}
% In this subsection, we show the detailed difference between DeepSAT and DG-DAGRNN. Or in Appendix.
% }

% \vspace{-8pt}
\section{Experiments}\label{sec:exp}
% \vspace{-4pt}

\begin{table*}[t!]
\centering
\caption{Performance comparison of DeepSAT and NeuroSAT on in-sample instances (\textit{Problems Solved}). %All models are trained on $230$k SR(3-10) training data.
We consider two settings: i) Same message passing iterations, where the number of message passing iterations is determined by the number of variables. ii) Test metric converges, where models iteratively run until no instance can be solved by increasing the number of iterations.} %The test sets follow the same formats used in training.}
% Format of \textit{Flat Circuit} refers to the circuit converted from CNF via Cube and Conquer~\cite{heule2011cube}.}
\label{tab:randomksat}
% \resizebox{\linewidth}{!}{%
\begin{tabular}{@{}ccccccccccccc@{}}
\toprule
                         &          & \multicolumn{5}{c}{Same Iterations}        &  & \multicolumn{5}{c}{Test Metric Converges}  \\ \midrule
Methods                  & Format   & SR($10$) & SR($20$) & SR($40$) & SR($60$) & SR($80$) &  & SR($10$) & SR($20$) & SR($40$) & SR($60$) & SR($80$) \\ \midrule
NeuroSAT                 & CNF      & 65\%   & 58\%   & 32\%   & 20\%   & 20\%   &  & 92\%   & 74\%   & 42\%   & 20\%   & 20\%   \\ \midrule
\multirow{2}{*}{DeepSAT} & Raw AIG  & 67\%   & 60\%   & 36\%   & 23\%   & 21\%   &  & 94\%   & 79\%   & 45\%   & 25\%   & 23\%   \\
                         & Opt. AIG & 72\%   & 66\%   & 40\%   & 31\%   & 23\%   &  & 98\%   & 85\%   & 51\%   & 37\%   & 26\%   \\ \bottomrule
\end{tabular}%
% }
% \vspace{-15pt}
\end{table*}

\subsection{Experimental settings} \label{Sec:Exp:Imp}
% \vspace{-4pt}

\textbf{Baselines.}
We compare our solution with the representative end-to-end method for SAT solving in the literature \footnote{We do not compare DeepSAT with DG-DAGRNN~\cite{amizadeh2018learning} as we cannot reproduce its results, despite that we have rigorously followed the descriptions in the paper.}%Our implementation of DG-DAGRNN is also attached in the supplementary materials.}
: NeuroSAT~\cite{selsam2018learning}.
NeuroSAT assumes that the input problem is in CNF. % while DG-DAGRNN uses a pre-processing procedure to convert the input CNFs into flat circuits~\cite{heule2011cube}.
There are several follow-up works of NeuroSAT~\cite{yolcu2019learning, kurin2020can, zhang2021nlocalsat, duan2022augment} that consuming CNFs. Nevertheless, they target on different settings and NeuroSAT still stands for a strong baseline for end-to-end SAT solving. %We give more discussion about the related models and implementation details in Appendix~\ref{appendix:baselines} and additional analysis in Appendix~\ref{appendix:additional}. 
To enable a fair comparison, we implement DeepSAT and NeuroSAT in a unified framework%\footnote{The code is available at: \url{https://anonymous.4open.science/r/deepsat_iclr23_E513/}.} 
with PyTorch Geometric (PyG)~\cite{pyg} package.

% Notably, we only report the comparison results with NeuroSAT in all experiments, even though DG-DAGRNN also targets on Circuit-SAT problems. We have tried our best to reproduce DG-DAGRNN and also applied the same EDA optimizations to it, as the official code of DG-DAGRNN remains publically unavailable up to now.  Unfortunately, with all the efforts, we still fail to train DG-DAGRNN to converge. We attribute the difficulty of training DG-DAGRNN to its vulnerability to training crash. On the one hand, DG-DARGNN employs a soft-evaluator as the reward network and is a variant of Policy Gradient method during training. As we observe empirically, the training is sensitive to several hyper-parameters and is extremely difficult to optimize even we follow the same setting described in the original paper. On the other hand, increasing the number of logical levels would result in a large stack of smoothed max and min functions in DG-DAGRNN. Consequently, the gradient tends to vanish and thus the training tends to fail. Moreover, even though we cannot reproduce DG-DAGRNN, the results DeepSAT obtains is still better than the ones reported in the original paper.

\textbf{Training datasets.}
We follow the SAT instance generation scheme proposed in NeuroSAT~\cite{selsam2018learning} to generative random $k$-SAT CNF pairs. We use SR($n$) to denote random $k$-SAT problems on $n$ variables generated by this scheme. In particular, we generate a SR($3-10$) dataset of $230$k SAT and UNSAT pairs. We restrict the scale of the training dataset due to limited GPU resources. However, we demonstrate that DeepSAT can generalize well to problems that are larger or with novel distributions, as shown below.
For NeuroSAT, these pairs are directly used for binary classification, while %DG-DAGRNN and 
DeepSAT is only trained on SAT instances. For DeepSAT, we convert CNFs to Raw AIG\footnote{Convert CNF to AIG using the cnf2aig tool:  \url{http://fmv.jku.at/cnf2aig/}}and Optimized AIG via logic synthesis (abbreviated as Opt. AIG). Note that in all experiments we convert the test instances to the same formats as the ones used in model training. % More details on different circuit representations are included in Appendix~\ref{appendix:circuit}.
% Moreover, we restrict the scale of the training dataset due to limited GPU resources. However, we demonstrate that DeepSAT can generalize well to problems that are larger or with novel distributions, as shown in the below experiments.

%Following \cite{amizadeh2018learning}, we generate $10K$ random instances of SR($3$)-SR($10$) in CNF. 
% It should be noted that the generation scheme proposed in ~\cite{selsam2018learning} will produce SAT instances with a relatively large number of clauses. There are few or even one valid assignment that can satisfy the instance. Therefore, the simulated probability is more likely to be a discrete number. To enable more effective supervision generation from logic simulation, we also produce $30K$ instances with fewer clauses, i.e., $3$ to $5$ times the number of variables. 
% After obtaining these 40K instances, we perform the logic synthesis process discussed in Section~\ref{subsec:transformation} to convert CNF instances into optimized AIGs. Then, we iteratively mask PO and some of PIs to calculate the simulated probability. % under $15K$ random patterns. 
% As a result, there are $280K$ AIG circuits for DeepSAT training. 

\textbf{Evaluation metric.}
DeepSAT is an \textit{incomplete}~\cite{zhang2021nlocalsat, amizadeh2018learning} solver, i.e., an instance is detected as satisfiable only after the model finds a satisfying assignment and otherwise returns \textit{unsolved}. 
Therefore, we only include satisfiable instances in the testing dataset for all experiments and measure the percentage of the successfully solved SAT instances out of all instances (abbreviated as \textit{Problems Solved}).  %DG-DAGRNN~\cite{amizadeh2018learning} also follows the same settings. 
% In Appendix~\ref{appendix:metric}, we discuss more details of the evaluation metrics for end-to-end SAT solvers.  

% \paragraph{Implementation details.}
% Our DeepSAT model has $69.13K$ trainable parameters. In particular, 
% DeepSAT consists of one forward propagation layer and one reverse propagation layer. The dimension of node hidden states is set as $64$. The regressor is a 3-layer MLP with hidden dimension $64$. 
% We train DeepSAT for $60$ epochs with batch-size $128$ on 4 Nvidia V100 GPUs. The Adam optimizer~\cite{kingma2014adam} is adopted with the learning rate $10^{-4}$ and weight decay $10^{-10}$. We use the topological batching technique introduced in \cite{thost2021directed} to accelerate the training.
% {\color{red}We provide the full details of model architecture and the ablation study of hyper-parameters in Appendix~\ref{appendix:model}. }

% \begin{figure}[t!]
% 	\centering
% 	\includegraphics[width=0.9\linewidth]{figs/Results.pdf}
% 	\vspace{-10pt}
% 	\caption{Performance comparison of DeepSAT, DG-DAGRNN and NeuroSAT}
% 	\label{FIG:result}
% % 	\vspace{-15pt}
% \end{figure}

% \vspace{-8pt}
\subsection{Random \textit{k}-SAT}
% \vspace{-4pt}

% This section compares the SAT solving performance on random $k$-SAT problem between our proposed DeepSAT and two baselines. 
This section shows comparisons of DeepSAT and NeuroSAT % and DG-DAGRNN, 
on solving random $k$-SAT problem. 
To evaluate the generalization performance, we validate our method on SAT instances generated from a much larger number of variables. In particular, %following the settings of \cite{amizadeh2018learning},
we evaluate the well-trained DeepSAT on SR($10$), SR($20$), SR($40$), SR($60$) and SR($80$), and compare with the results with the baseline.

% Since the assignment decoding schemes of DeepSAT and baselines are different, we conduct two experiments for performance comparison. As DeepSAT assigns one logic value for a single PI in one bidirectional message passing iteration (i.e., one forward propagation and one reverse propagation), it needs $I$ iterations of bidirectional message passing in total to obtain an assignment for a $I$-variable SAT instance. NeuroSAT and DG-DAGRNN have the similar bidirectional message passing scheme, yet they generate the assignment for a specific problems \textit{once} after a number of bidirectional message passing iterations. Therefore, for a $I$-variable SAT instance, we fix the number of message passing iterations as $I$, in which case DeepSAT can generate one and only one complete assignment and baselines are aligned with similar message passing cost. Following \cite{amizadeh2018learning}, the second setting we consider is that we let models iteratively run on test dataset until test metric \textit{Problems Solved} converges. It should noted that under this setting NeuroSAT and DG-DAGRNN are queried for multiple times for assignment decoding under different iterations of message passing, in order to monitor the test metric. In other words, both baselines will generate multiple assignments for one SAT instance. As for DeepSAT, we follow the proposed solution sampling scheme which samples $I+1$ solutions in worst case. In this case, we compare the upper-bound of performance regardless of computational cost.

Since DeepSAT and NeuroSAT use different assignment decoding schemes to yield the SAT assignments given an SAT instance, 
% and more assignments are usually accompanied by more computational cost, but lead to higher successful rate, 
we consider two experimental settings for performance comparison: i) compare under the same computational cost, which is measured by the number of message passing iterations, and ii) compare the upper-bound of performance regardless of the computational cost. 
For the first setting, we fix the number of message passing iterations as $I$ for an $I$-variable SAT instance. This is because DeepSAT needs to perform at least $I$ message passing iterations to generate a solution for an $I$-variable SAT instance, while NeuroSAT can generate the solution for a specific problem \textit{once} by clustering on the literal embedding after a number of bidirectional message passing iterations. Hence, for a $I$-variable SAT instance, we fix the number of message passing iterations as $I$, in which case DeepSAT can generate one and only one complete assignment and NeuroSAT are aligned with same message passing cost.
For the second setting, we let both models generate as many assignments as possible until the test metric converges. In this way, NeuroSAT is queried for multiple times for assignment decoding under different iterations of message passing, in order to monitor the test metric. In other words, NeuroSAT will generate multiple assignments for one SAT instance. As for DeepSAT, we follow the proposed solution sampling scheme introduced in Section~\ref{subsec:sat-infer}, which samples at most $I+1$ solutions in worst case.

% {
% \color{blue}
% \paragraph{Comparison under same message passing iterations.}
% Here we consider the case that all models run the same iterations.

% \paragraph{Comparison when test metric converges.}
% Here, we consider the case where models iteratively run on each test dataset until the test metric converges.
% }

We present \textit{Problems Solved} on different testing datasets with different formats in Table~\ref{tab:randomksat}.
%\footnote{\color{red}From DG-DAGRNN: We let both models iteratively run on each test dataset until the test metric converges. In other words, they sample more than one solution per problem.}. 
From this table, we have several observations. First, compared with the baselines, DeepSAT can achieve the highest \textit{Problems Solved} on all evaluated datasets under both experiment settings, with a significant margin. For example, on SR($20$), DeepSAT trained on Opt. AIGs can solve $85\%$ of problems, while NeuroSAT trained on CNFs only solve $74\%$ of problems. 
% \footnote{Since DG-DAGRNN is not open-sourced and we cannot reproduce}. 
Second, the performance of both models degrades as we increase the number of variables. Yet, DeepSAT generalizes better to bigger and harder problems than NeuroSAT. %across all circuit formats. 
% For instance, DeepSAT trained on Opt. AIGs can solve $7.69\%$ more problems on SR($10$) dataset than NeuroSAT, while the relative improvement increases to $71.43\%$ on SR($80$) dataset.
% More precisely, when the problem enlarges from SR($3)-SR($10$) to SR($80$), the problem solved of NeuroSAT decreases $90.22\%$ (from $92\%$ to $9\%$) and DG-DAGRNN decreases $84.62\%$ (from $91\%$ to $14\%$), while our model shows only $75.51\%$ performance degrade (from $98\%$ to $24\%$).

To further validate the effectiveness of our proposed formulation, i.e., SAT solving as a generative modeling procedure, we monitor \textit{Problems Solved} during the solution sampling procedure on SR($10$). By sampling a single solution for each problem, DeepSAT can solve $72\%$ of all problems. If we sample two more solutions for each problem from DeepSAT, the percentage of solved problems increases to $93\%$. Contrarily, to solve $92\%$ of problems on SR($10$), NeuroSAT %and DG-DAGNN 
requires additional tens of iterations of message propagation to make prediction coverage. 
DeepSAT terminates when the latest sampled solutions are satisfying and samples $1.63$ solutions on average for SR($10$). The result on other test datasets shows the same tendency.
% See Appendix~\ref{appendix:sample} for more results on the proposed solution sampling scheme.
% To conclude, with a novel formulation, DeepSAT is more efficient than the baselines.

% \begin{figure}[t!]
% 	\centering
% 	\includegraphics[width=0.7\linewidth]{figs/Results.pdf}
% 	\caption{Performance comparison of DeepSAT, DG-DAGRNN~\cite{amizadeh2018learning} and NeuroSAT~\cite{selsam2018learning}{\color{red}Modify this figure.}}
% 	\label{FIG:result}
% \end{figure}

% \begin{table}[t!]
% \caption{The number of problems solved v.s. the number of sampled solutions for SR(10)}\label{tab:sr10}
% \centering
% \begin{tabular}{@{}cccccccccccc@{}}
% \toprule
% \# Sample Solutions & 1  & 2 & 3  & 4 & 5 & 6 & 7 & 8 & 9 & 10 & 11 \\ \midrule
% \# Problem Solved   & 70 & 8 & 13 & 1 & 2 & 2 & 1 & 0 & 1 & 0  & 0  \\ \bottomrule
% \end{tabular}
% \end{table}

\vspace{-2pt}
\subsection{Effectiveness of Logic Synthesis}
% \vspace{-4pt}

In Table~\ref{tab:randomksat}, we see that DeepSAT maintains state-of-the-art performance even when trained on raw AIGs without logic synthesis. Specifically, DeepSAT trained on raw AIGs outperforms NeuroSAT trained on CNFs consistently across all test sets, despite that raw AIGs include less structural information for learning.  For example, DeepSAT trained on raw AIGs successfully solves $94$ and $79$ out of $100$ SR($10$) and SR($20$) problems, respectively, which is better % than both DG-DAGRNN trained on Optimized AIGs and 
NeuroSAT trained on CNFs.
% Moreover, the problem solved percentage of w/ syn model is $18\%$ more than w/o syn model in the SR($3$)-SR($10$) testing dataset. For SR($20$), the w/ syn model can achieve $85\%$ \textit{Problems Solved}, even twice than the w/o syn model. % To conclude, logic synthesis enables better generalization ability of DeepSAT model. 
In summary, we believe that the performance gain of DeepSAT not only comes from EDA optimization, but also thanks to the customized bidirectional propagation with polarity prototypes that can effectively model the conditional distribution of possible assignments.

\begin{table}[t!]
\caption{The comparison of DeepSAT and NeuroSAT on novel distributions.}\label{tab:novel-dist}
\centering
\resizebox{\linewidth}{!}{%
\begin{tabular}{@{}ccccccc@{}}
\toprule
Method                   & Format   & Coloring Acc. & Domset Acc. & Clique Acc. & Vertex Acc. & Avg.  Acc. \\ \midrule
NeuroSAT                 & CNF      & 0\%           & 44\%        & 35\%        & 0\%         & 22\%       \\ \midrule
\multirow{2}{*}{DeepSAT} & Raw AIG  & 63\%          & 81\%        & 77\%        & 82\%        & 76\%       \\
                         & Opt. AIG & 98\%          & 99\%        & 92\%        & 97\%        & 97\%       \\ \bottomrule
\end{tabular}%
}
\vspace{-15pt}
\end{table}

\vspace{-2pt}
\subsection{Novel Distributions}
% \vspace{-4pt}

To further investigate the generalization ability of our DeepSAT model, we evaluate the model trained on SR($3-10$) on the different NP-Complete problems with novel distributions: graph $k$-coloring, vertex $k$-cover, $k$-clique-detection and dominating $k$-set. The above problems can be reduced to the SAT problems and solved by a SAT solver. We generate 100 random graphs with $6-10$ nodes and the edge percentage of $37\%$ for each problem. For each graph in each problem type, we generate graph coloring instances ($3 \leq k \leq 5$), dominating set problems ($2 \leq k \leq 4$), clique-detection problems ($3 \leq k \leq 5$) and vertex cover problems ($4 \leq k \leq 6$). We report the results when the test metric converges.
Notably, the generated problem instances are not particularly difficult for state-of-the-art SAT solvers. 
% However, both baselines~\cite{amizadeh2018learning, selsam2018learning} report relatively poor results on these novel instances. 
% For example, DG-DAGRNN can only solve half of graph $k$-coloring problems. 
For our purpose, the evaluated results in this section serve as simple benchmarks to demonstrate the superiority of DeepSAT over the baseline. 

As shown in Table~\ref{tab:novel-dist}, DeepSAT trained on Opt. AIGs can solve $98\%$ of graph coloring problems, $99\%$ of dominating set problems, $92\%$ of clique-detection problems, and $97\%$ of vertex cover problems. When DeepSAT is trained on raw AIGs, the performance degrades to some extent, yet is still better than NeuroSAT on these datasets.
In contrast, we observe there is an apparent discrepancy in performance of NeuroSAT between novel distribution problems and the in-sample distribution used for training, even though the size of problems is similar. To sum up, DeepSAT maintains the same solving ability on novel problems, and the pre-processing of logic synthesis enables better generalization ability of novel distributions.
% We conjecture this is due to the novel bidirectional propagation with polarity prototypes, which mimics 

% We compare our model with DG-DAGRNN~\cite{amizadeh2018learning} on solving graph coloring problems. As claimed by the authors, DG-DAGRNN only finds a solution for $48\%$ out of all problems, far from $91\%$ problem solved of the random $k$-SAT problem. Meanwhile, we compare DeepSAT with NeuroSAT~\cite{selsam2018learning} on the other novel distribution problems. The authors claim their model can successfully solve $85\%$ problems in the mixed dataset. The NeuroSAT is still unstable because there is an apparent discrepancy in performance between novel distribution problems and the similar distribution problems as the training dataset ($70\%$ random $k$-SAT problems solved). Contrarily, since we uniform the distribution by logic synthesis, DeepSAT is not sensitive to the diversity of problem instances. 

% \subsection{\color{red}Logic Equivalent Checking}
% {
% \color{red}
% \textbf{Train our model and baselines on LEC. We expect LEC benchmarks to include more structure information, and DeepSAT can perform better in this case.}
% }

% \section{Discussion}\label{sec:discussion}

% For the regularized latent space, we may discuss a little bit about the interpolation of representation?

% It should be regarded as one of methods that transferring the knowledge from EDA field into machine learning.

% Also benefit the SAT problems in circuit context. For example, SAT-based ATPG, no need to do the Tseitin transformation for every fault.

% \vspace{-8pt}
\section{Conclusion and Future Work}\label{sec:conclusion}
% \vspace{-4pt}

This paper presents \textit{DeepSAT}, an EDA-driven end-to-end learning framework for SAT solving. Specifically, we first propose to apply a logic synthesis-based pre-processing procedure to reduce the distribution diversity among various SAT instances. Next, we formulate SAT solving as a conditional generative procedure and construct more informative supervisions with simulated logic probabilities. A dedicated directed acyclic graph neural network with polarity prototypes is then trained to approximate the conditional distribution of SAT solutions. Our experimental results show that DeepSAT outperforms the existing end-to-end baseline. 

While DeepSAT demonstrates considerable improvements in end-to-end SAT solving, there is still a significant performance gap compared to state-of-the-art heuristic-based SAT solvers. In our future work, on the one hand, we plan to improve its performance further; on the other hand, we would also explore novel joint solutions that combine them, e.g., using constraint propagation mechanism learned in DeepSAT to guide better heuristic in classical Circuit-SAT solvers.

%Although the performance is still inferior than classical SAT solvers, we view DeepSAT as a step towards practical, scalable, and accurate end-to-end solutions for SAT.


\balance
\bibliographystyle{IEEEtran}
\bibliography{reference}

\begin{thebibliography}{10}
\providecommand{\url}[1]{#1}
\csname url@samestyle\endcsname
\providecommand{\newblock}{\relax}
\providecommand{\bibinfo}[2]{#2}
\providecommand{\BIBentrySTDinterwordspacing}{\spaceskip=0pt\relax}
\providecommand{\BIBentryALTinterwordstretchfactor}{4}
\providecommand{\BIBentryALTinterwordspacing}{\spaceskip=\fontdimen2\font plus
\BIBentryALTinterwordstretchfactor\fontdimen3\font minus
  \fontdimen4\font\relax}
\providecommand{\BIBforeignlanguage}[2]{{%
\expandafter\ifx\csname l@#1\endcsname\relax
\typeout{** WARNING: IEEEtran.bst: No hyphenation pattern has been}%
\typeout{** loaded for the language `#1'. Using the pattern for}%
\typeout{** the default language instead.}%
\else
\language=\csname l@#1\endcsname
\fi
#2}}
\providecommand{\BIBdecl}{\relax}
\BIBdecl

\bibitem{buttner2005satisfiability}
M.~B{\"u}ttner and J.~Rintanen, ``Satisfiability planning with constraints on
  the number of actions.'' in \emph{International Conference on Automated
  Planning and Scheduling}, 2005, pp. 292--299.

\bibitem{horbach2010boolean}
A.~Horbach, ``A boolean satisfiability approach to the resource-constrained
  project scheduling problem,'' \emph{Annals of Operations Research}, vol. 181,
  no.~1, pp. 89--107, 2010.

\bibitem{vizel2015boolean}
Y.~Vizel, G.~Weissenbacher, and S.~Malik, ``Boolean satisfiability solvers and
  their applications in model checking,'' \emph{Proceedings of the IEEE}, vol.
  103, no.~11, pp. 2021--2035, 2015.

\bibitem{sorensson2005minisat}
N.~Sorensson and N.~Een, ``Minisat v1. 13-a sat solver with conflict-clause
  minimization,'' \emph{SAT}, vol. 2005, no.~53, pp. 1--2, 2005.

\bibitem{fleury2020cadical}
A.~B. K. F.~M. Fleury and M.~Heisinger, ``Cadical, kissat, paracooba,
  plingeling and treengeling entering the sat competition 2020,'' \emph{SAT
  COMPETITION}, vol. 2020, p.~50, 2020.

\bibitem{kurin2020can}
V.~Kurin, S.~Godil, S.~Whiteson, and B.~Catanzaro, ``Can q-learning with graph
  networks learn a generalizable branching heuristic for a sat solver?''
  \emph{Advances in Neural Information Processing Systems}, vol.~33, pp.
  9608--9621, 2020.

\bibitem{yolcu2019learning}
E.~Yolcu and B.~P{\'o}czos, ``Learning local search heuristics for boolean
  satisfiability,'' \emph{Advances in Neural Information Processing Systems},
  vol.~32, 2019.

\bibitem{zhang2021nlocalsat}
W.~Zhang, Z.~Sun, Q.~Zhu, G.~Li, S.~Cai, Y.~Xiong, and L.~Zhang, ``Nlocalsat:
  boosting local search with solution prediction,'' in \emph{Proceedings of the
  Twenty-Ninth International Conference on International Joint Conferences on
  Artificial Intelligence}, 2021, pp. 1177--1183.

\bibitem{selsam2018learning}
D.~Selsam, M.~Lamm, B.~B\"{u}nz, P.~Liang, L.~de~Moura, and D.~L. Dill,
  ``Learning a {SAT} solver from single-bit supervision,'' in
  \emph{International Conference on Learning Representations}, 2019.

\bibitem{amizadeh2018learning}
S.~Amizadeh, S.~Matusevych, and M.~Weimer, ``Learning to solve circuit-{SAT}:
  An unsupervised differentiable approach,'' in \emph{International Conference
  on Learning Representations}, 2019.

\bibitem{duan2022augment}
H.~Duan, P.~Vaezipoor, M.~B. Paulus, Y.~Ruan, and C.~Maddison, ``Augment with
  care: Contrastive learning for combinatorial problems,'' in
  \emph{International Conference on Machine Learning}.\hskip 1em plus 0.5em
  minus 0.4em\relax PMLR, 2022, pp. 5627--5642.

\bibitem{wu2007qutesat}
C.-A. Wu, T.-H. Lin, C.-C. Lee, and C.-Y. Huang, ``Qutesat: a robust
  circuit-based sat solver for complex circuit structure,'' in \emph{2007
  Design, Automation \& Test in Europe Conference \& Exhibition}.\hskip 1em
  plus 0.5em minus 0.4em\relax IEEE, 2007, pp. 1--6.

\bibitem{bjesse2004dag}
P.~Bjesse and A.~Boralv, ``Dag-aware circuit compression for formal
  verification,'' in \emph{IEEE/ACM International Conference on Computer Aided
  Design, 2004. ICCAD-2004.}\hskip 1em plus 0.5em minus 0.4em\relax IEEE, 2004,
  pp. 42--49.

\bibitem{mishchenko2006dag}
A.~Mishchenko, S.~Chatterjee, and R.~Brayton, ``Dag-aware aig rewriting: A
  fresh look at combinational logic synthesis,'' in \emph{2006 43rd ACM/IEEE
  Design Automation Conference}.\hskip 1em plus 0.5em minus 0.4em\relax IEEE,
  2006, pp. 532--535.

\bibitem{oord2016conditional}
A.~v.~d. Oord, N.~Kalchbrenner, O.~Vinyals, L.~Espeholt, A.~Graves, and
  K.~Kavukcuoglu, ``Conditional image generation with pixelcnn decoders,'' in
  \emph{Proceedings of the 30th International Conference on Neural Information
  Processing Systems}, 2016, pp. 4797--4805.

\bibitem{khalil2017learning}
E.~Khalil, H.~Dai, Y.~Zhang, B.~Dilkina, and L.~Song, ``Learning combinatorial
  optimization algorithms over graphs,'' \emph{Advances in neural information
  processing systems}, vol.~30, 2017.

\bibitem{hudson2021graph}
B.~Hudson, Q.~Li, M.~Malencia, and A.~Prorok, ``Graph neural network guided
  local search for the traveling salesperson problem,'' in \emph{International
  Conference on Learning Representations}, 2022.

\bibitem{wu2020comprehensive}
Z.~Wu, S.~Pan, F.~Chen, G.~Long, C.~Zhang, and S.~Y. Philip, ``A comprehensive
  survey on graph neural networks,'' \emph{IEEE transactions on neural networks
  and learning systems}, vol.~32, no.~1, pp. 4--24, 2020.

\bibitem{thost2021directed}
V.~Thost and J.~Chen, ``Directed acyclic graph neural networks,'' in
  \emph{International Conference on Learning Representations}, 2021.

\bibitem{li2021representation}
M.~Li, S.~Khan, Z.~Shi, N.~Wang, Y.~Huang, and Q.~Xu, ``Deepgate: learning
  neural representations of logic gates,'' in \emph{Proceedings of the 59th
  ACM/IEEE Design Automation Conference}, 2022, pp. 667--672.

\bibitem{cortadella2003timing}
J.~Cortadella, ``Timing-driven logic bi-decomposition,'' \emph{IEEE
  Transactions on Computer-Aided Design of Integrated Circuits and Systems},
  vol.~22, no.~6, pp. 675--685, 2003.

\bibitem{brayton2010abc}
R.~Brayton and A.~Mishchenko, ``Abc: An academic industrial-strength
  verification tool,'' in \emph{International Conference on Computer Aided
  Verification}.\hskip 1em plus 0.5em minus 0.4em\relax Springer, 2010, pp.
  24--40.

\bibitem{walker1976locally}
A.~Walker and D.~Wood, ``Locally balanced binary trees,'' \emph{The Computer
  Journal}, vol.~19, no.~4, pp. 322--325, 1976.

\bibitem{toda2016implementing}
T.~Toda and T.~Soh, ``Implementing efficient all solutions sat solvers,''
  \emph{Journal of Experimental Algorithmics (JEA)}, vol.~21, pp. 1--44, 2016.

\bibitem{snell2017prototypical}
J.~Snell, K.~Swersky, and R.~Zemel, ``Prototypical networks for few-shot
  learning,'' in \emph{Proceedings of the 31st International Conference on
  Neural Information Processing Systems}, 2017, pp. 4080--4090.

\bibitem{belov2010improved}
A.~Belov and Z.~Stachniak, ``Improved local search for circuit
  satisfiability,'' in \emph{International Conference on Theory and
  Applications of Satisfiability Testing}.\hskip 1em plus 0.5em minus
  0.4em\relax Springer, 2010, pp. 293--299.

\bibitem{zhang2021circuit}
H.-T. Zhang, J.-H.~R. Jiang, and A.~Mishchenko, ``A circuit-based sat solver
  for logic synthesis,'' in \emph{2021 IEEE/ACM International Conference On
  Computer Aided Design (ICCAD)}.\hskip 1em plus 0.5em minus 0.4em\relax IEEE,
  2021, pp. 1--6.

\bibitem{pyg}
M.~Fey and J.~E. Lenssen, ``Fast graph representation learning with {PyTorch
  Geometric},'' in \emph{ICLR Workshop on Representation Learning on Graphs and
  Manifolds}, 2019.

\end{thebibliography}

% \newpage
% \appendix
% \input{appendix}

\end{document}